%% file: 0_main.tex
\documentclass[letterpaper, 10 pt, conference]{ieeeconf}  % Comment this line out if you need a4paper

\IEEEoverridecommandlockouts                              % This command is only needed if 
\usepackage{lipsum} 
\usepackage{epsfig}
\usepackage{float}
\usepackage{bbold}
\usepackage{bm}
\usepackage{hyperref}
\usepackage{hhline} % required for double horiz line in tables
\usepackage{makecell} % required for break line in cells
\usepackage{pifont}  % required for validation and error symbols in tables
\usepackage[noadjust]{cite}

\usepackage{color,soul} % added for yellow highlighting

\usepackage{amsmath}

\title{\LARGE \bf
QDGset: A Large Scale Grasping Dataset Generated with Quality-Diversity
}

\usepackage[dvipsnames]{xcolor} %xcolor is REQUIRED to define color in \newcommand

\newif\ifJoh
\newif\ifMat
\newif\ifFra
\newif\ifIgn

\Johtrue
\Mattrue
\Fratrue
\Igntrue

\begin{document}

\author{Johann Huber$^{*1}$, François Hélénon$^{*1}$, Mathilde Kappel$^{*1}$, Ignacio de Loyola Páez-Ubieta$^{*2}$, \\
Santiago T. Puente$^{2}$, Pablo Gil$^{2}$, Faïz Ben Amar$^{1}$ and Stéphane Doncieux$^{1}$ 
\thanks{$^*$ equal contribution and corresponding authors}
\thanks{$^{1}$Sorbonne Université, CNRS, Institut des Systèmes Intelligents et de Robotique, ISIR, F-75005 Paris, France {\tt\small \{huber, helenon, kappel, benamar, doncieux\}@isir.upmc.fr}}
\thanks{$^{2}$University of Alicante, University Institute for Computing Research (IUII), AUtomatics, RObotics and Artificial Vision research group (AUROVA), E-03690 San Vicente del Raspeig (Alicante), Spain {\tt\small \{ignacio.paez, santiago.puente, pablo.gil\}@ua.es}}}

\maketitle
\thispagestyle{empty}
\pagestyle{empty}

%\renewcommand{\thefootnote}{\fnsymbol{footnote}}
%\footnote[1]{text}

%%%%%%%%%%%%%%%%%%%%%%%%%%%%%%%%%%%%%%%%%%%%%%%%%%%%%%%%%%%%%%%%%%%%%%%%%%%%%%%%

%%%%%%%%%%%%%%%%%%%%%%%%%%%%%%%%%%%%%%%%%%%%%%%%%%%%%%%%%%%%%%%%%%%%%%%%%%%%%%%%
%                                 Paper's body
%%%%%%%%%%%%%%%%%%%%%%%%%%%%%%%%%%%%%%%%%%%%%%%%%%%%%%%%%%%%%%%%%%%%%%%%%%%%%%%%

\input{tex_files/abstract}

\input{tex_files/1_introduction}

\input{tex_files/2_related_works}

\input{tex_files/3_method}

\input{tex_files/4_experiments}

\input{tex_files/5_results_and_discussion}

\input{tex_files/6_conclusions}

\input{tex_files/acknowledgment}


%%%%%%%%%%%%%%%%%%%%%%%%%%%%%%%%%%%%%%%%%%%%%%%%%%%%%%%%%%%%%%%%%%%%%%%%%%%%%%%%

\bibliographystyle{IEEEtran}

\input{tex_files/bilbio}
\end{document}

%% file: tex_files/abstract.tex
%%%%%%%%%%%%%%%%%%%%%%%%%%%%%%%%%%%%%%%%%%%%%%%%%%%%%%%%%%%%%%%%%%%%%%%
%                            Abstract
%%%%%%%%%%%%%%%%%%%%%%%%%%%%%%%%%%%%%%%%%%%%%%%%%%%%%%%%%%%%%%%%%%%%%%%

\begin{abstract}

Recent advances in AI have led to significant results in robotic learning, but skills like grasping remain partially solved. Many recent works exploit synthetic grasping datasets to learn to grasp unknown objects. However, those datasets were generated using simple grasp sampling methods using priors. Recently, Quality-Diversity (QD) algorithms have been proven to make grasp sampling significantly more efficient. In this work, we extend QDG-6DoF, a QD framework for generating object-centric grasps, to scale up the production of synthetic grasping datasets. We propose a data augmentation method that combines the transformation of object meshes with transfer learning from previous grasping repertoires. The conducted experiments show that this approach reduces the number of required evaluations per discovered robust grasp by up to 20\%. We used this approach to generate QDGset, a dataset of 6DoF grasp poses that contains about 3.5 and 4.5 times more grasps and objects, respectively, than the previous state-of-the-art. Our method allows anyone to easily generate data, eventually contributing to a large-scale collaborative dataset of synthetic grasps. 

%IGN: Supervisor tells he is not able to understand highlighted sentence. Maybe rephrasing it...(?) 

%J: the simple sampling methods are extensively studied in Eppner et al, 2022 - and we showed in our IROS'24 paper that QDG-6DoF was much more efficient for sampling grasps poses on a known 3d model than those ones.
%
% I guess this sentence is clearer when knowing more about those methods. I don't know how could it be more clear, while not going into too much details - as it is an abstract ...
%
% Eppner, C., Mousavian, A. \& Fox, D. (2022). A billion ways to grasp: An evaluation of grasp sampling schemes on a dense, physics-based grasp data set. In 2022 International Symposium of Robotics Research (ISRR), 20 (pp. 890-905). Springer. doi: 10.1007/978-3-030-95459-8\_55.
%
% Huber, J., Hélénon, F., Kappel, M., Chelly, E., Khoramshahi, M., Amar, F. B. \& Doncieux, S. (2024). Speeding up 6-DoF Grasp Sampling with Quality-Diversity. In 2024 IEEE/RSJ International Conference on Intelligent Robots and Systems (IROS). doi: 10.48550/arXiv.2403.06173

\end{abstract}

%% file: tex_files/1_introduction.tex
%%%%%%%%%%%%%%%%%%%%%%%%%%%%%%%%%%%%%%%%%%%%%%%%%%%%%%%%%%%%%%%%%%%%%%%
%                           Introduction
%%%%%%%%%%%%%%%%%%%%%%%%%%%%%%%%%%%%%%%%%%%%%%%%%%%%%%%%%%%%%%%%%%%%%%%

%The purpose of this paper is not to provide another fixed dataset, but to provide a methodology to easily generate grasping data.

\section{INTRODUCTION}

Grasping is a critical skill in robotics, serving as a prerequisite for numerous manipulation tasks. The previously dominant analytical-based methods \cite{nguyen1988constructing} have been increasingly overtaken by data-driven strategies since the early 21st century \cite{zhang2022robotic}. However, the challenging exploration aspect of grasping prevents the initialization of the learning process, as most grasps attempted by a randomly initialized policy yield no reward \cite{huber2023quality}. Imitation learning \cite{qin2022from,wang2021demograsp,sefat2022singledemograsp}, parallel grippers \cite{depierre2018jacquard,fang2020graspnet} and top-down grasps \cite{levine2018handeye,mahler2017dexnet2,yang2023pave} are the most frequently employed solutions to address this challenge. However, these methods restrict the operational space, consequently limiting the flexibility and adaptability of the resulting policies.

Recent advances in robotic learning highlight the effectiveness of data-driven approaches for skill acquisition \cite{chi2023diffusion,octo2023octo,urain2023se3diffusionfield,barad2023graspldm,chen2024nsgf}. These results depend on advanced artificial intelligence techniques that necessitate vast amounts of high-quality data to generalize effectively to new scenarios. This has resulted in the release of large datasets \cite{padalkar2023openxembodiement}, many of which focus on grasping \cite{fang2020graspnet,eppner2021acronym,turpin2023fastgraspd}. Acquiring high-quality datasets is increasingly crucial for enabling high-representational-power architectures to achieve generalization \cite{chi2023diffusion} and for developing efficient foundation models \cite{octo2023octo} that can transfer skills across platforms and scenarios.

%IGN: supervisor says there are many cites without detail. Suggests removing 2 or 3 less relevant cites.

% J: It aims to give an overview of the current SOTA, like literally the past year and a half. Citing that means : here is the current trend from a high level point of view, take a look a it if you want. Now the key idea is that we need DATA, and that is exactly the topic of the present paper. Hoora!

Many grasping datasets can be annotated and collected from real world \cite{lenz2015deep, chu2018real,levine2018handeye,pinto2016supersizing}, but this approach cannot be easily scaled to reach the required size to get the most out of the data-driven methods. A recent trend consists of automatically generated grasping datasets using simulations to validate the grasps \cite{eppner2021acronym, eppner2023abw2g, gilles2022metagraspnet}. The most successful one is ACRONYM \cite{eppner2021acronym}, which has recently been used for training numerous advanced AI methods \cite{chi2023diffusion, barad2023graspldm, chen2024nsgf}.

Quality-Diversity (QD) is a family of algorithms dedicated to generating diverse and high-performing solutions to given problems \cite{cully2022qd}. QDG-6DoF \cite{huber2024speeding} is a recently released framework that leverages QD and robotic priors to produce grasping datasets. This framework demonstrated to be at least twice more sample efficient than the state-of-the-art methods used by ACRONYM to produce the grasps.  

%IGN: it is a general affirmation, ypu should detail a bit more

% J: The citation gives a comprehensive overview of what "a given problem" means. What matter is the grasping part, which is detailed right after.

\input{tex_files/figures/vis_intro_qdgset}

This work extends QDG-6DoF to scale up the generation of synthetic grasping datasets. In particular:

\begin{itemize}
    \item We extend QDG-6DOF to data augmentation by combining the transformations of object meshes with a QD transfer learning approach;
    \item An experiment conducted on 450 augmented objects shows that the required number of evaluations to generate a robust grasp is reduced by up to 20\%, compared to the current best grasp sampling method;
    \item We leverage this approach to introduce \textit{QDGset} (Fig. \ref{fig:vis_intro_qdgset}), a large-scale dataset of about 60M parallel jaw gripper grasps and 40k simulated objects, which is significantly larger than the state-of-the-art;
\end{itemize}

This paper does not aim to share a new static grasping dataset but rather proposes a method for easily generating data for sim2real purposes. The code and dataset had thus been made publicly available on the project website\footnote{\url{https://qdgrasp.github.io/}}, as well as how to contribute to the QDGset initiative.

%% file: tex_files/figures/vis_intro_qdgset.tex
\begin{figure}[t]
  \centering
\centering
  \includegraphics[width=0.8\columnwidth]{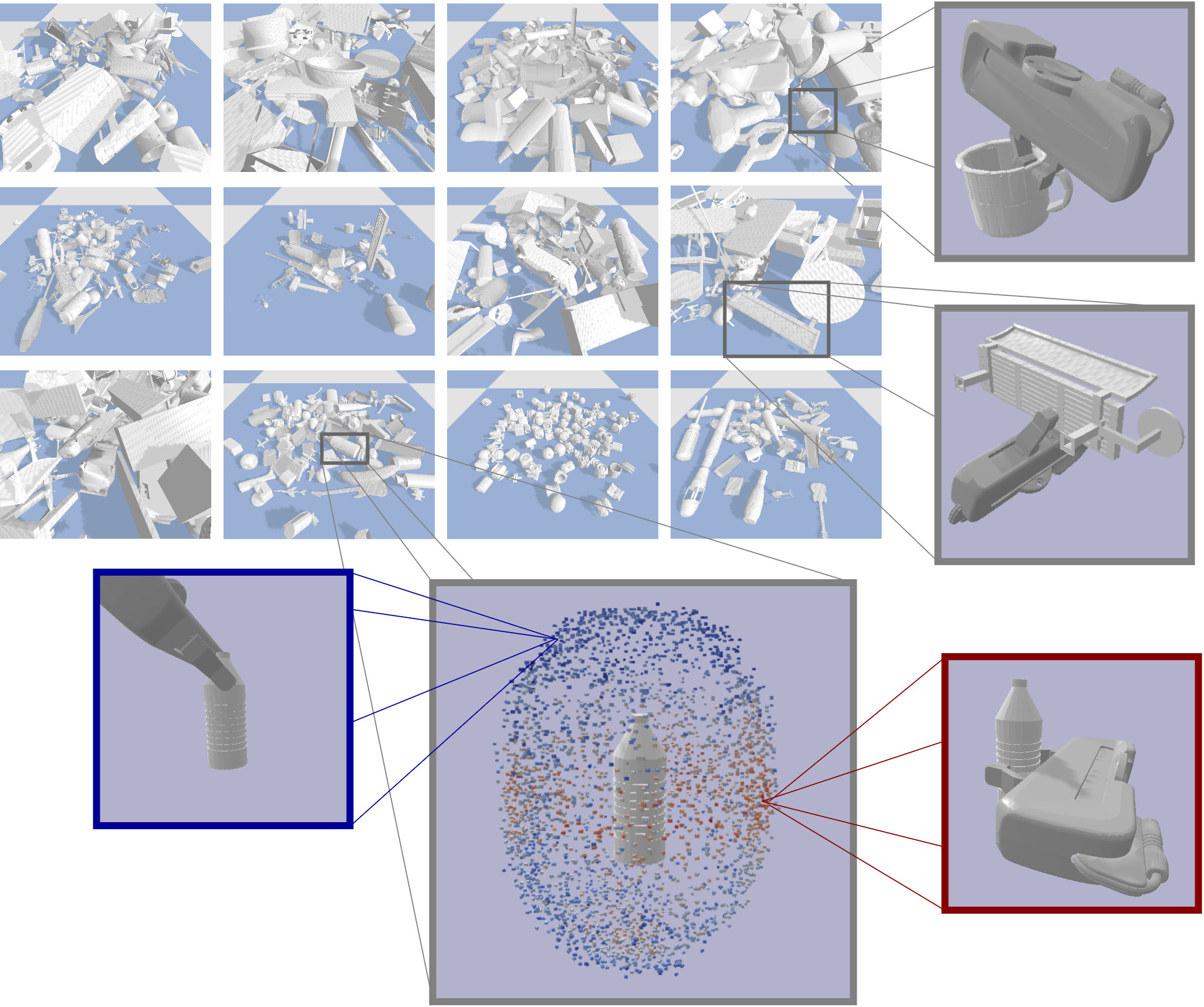}
  \caption{QDGset contains 62M 6DoF grasps on about 40k simulated objects. Each grasp is labelled with a probability to transfer in the real world \cite{huber2023domainrandomization}.}
  \label{fig:vis_intro_qdgset}
\end{figure}

%objects have been randomly sampled from some of the object datasets included in QDG-Set, including gripper scale \hl{()}, human-body scale \hl{()}, and adversarial \hl{()} objects. Each voxel corresponds to the end effector position of a successful grasp pose. The color describes the probability to transfer into the real world \cite{huber2023domainrandomization}, the hotter the higher.

%% file: tex_files/2_related_works.tex
%%%%%%%%%%%%%%%%%%%%%%%%%%%%%%%%%%%%%%%%%%%%%%%%%%%%%%%%%%%%%%%%%%%%%%%
%                           Related works
%%%%%%%%%%%%%%%%%%%%%%%%%%%%%%%%%%%%%%%%%%%%%%%%%%%%%%%%%%%%%%%%%%%%%%%

\section{RELATED WORKS}
\label{sec:2_related_works}

%IGN: in general, i dont know if you can make a table, scheme or figure to show the taxonomy from state of the art works

% J : Ideally yes, but we have 6 pages. So we'll have to stick to that ... the database table plays this role

%=====================================================================%
%                   2.1) Datasets for grasping in robotics
%=====================================================================%

%--------------------------------------------%
% Learning to Grasp in Robotics
%--------------------------------------------%

% todo rephrase
\textbf{\textit{Learning to Grasp in Robotics.}} Since the rise of data-driven methods in the field, numerous paradigms have been investigated, such as reinforcement learning \cite{chen2023rlgrasp1,zhou2023rlgrasp3} and learning from a limited number of demonstrations \cite{wang2021demograsp,sefat2022singledemograsp}. However, to make these methods feasible for exploration and learning, it is necessary to constrain the search space due to the rarity of successful grasps among all potential grasps. Numerous data-greedy approaches have been proposed to achieve generalization capabilities \cite{levine2018handeye,mahler2017dexnet2,fang2020graspnet}. Recently were used Diffusion models \cite{urain2023se3diffusionfield}, Variational autoencoders \cite{barad2023graspldm}, or 
neural representations \cite{chen2024nsgf} to learn to grasp unknown objects. It is worth noting that the majority of these approaches leverage automatically generated data to train their models.

%IGN: from a computational viewpoint in the training phase? from a viewpoint of finding a suboptimal grasp? It should be clarified

% J: slightly clarified, I go back to it in the end.

%IGN: the other supervisor suggests removing this last sentence "Interestingly, most of them exploit automatically generated data in their learning process"

% J: No way, it is the main idea of the paper. SOTA exploit a dataset calls ACRONYM, and we do better. Removing that literally means removing the core idea of the paper.

%--------------------------------------------%
% Grasping Datasets
%--------------------------------------------%
 
\textbf{\textit{Grasping Datasets.}} The breakthrough implied by large datasets in image processing \cite{deng2009imagenet} incites similar initiatives for grasping. The Columbia Grasp dataset \cite{goldfeder2009columbia} aims to "\textit{collect enough grasping data to build new grasp planners based on learning}." Numerous paradigms were explored to that end. Self-supervision \cite{levine2018handeye,pinto2016supersizing} or manual annotations \cite{lenz2015deep, chu2018real} showed promising results but are not easily scalable. Automatic collection or annotation processes were proposed to circumvent this problem. Fang et al. collected data from real cluttered scenes and annotated grasp poses based on analytic criteria but are still limited by real data collection \cite{fang2020graspnet}. An increasing number of works relies on a simulated interaction between an object and a gripper to label the grasps \cite{eppner2021acronym, eppner2023abw2g, gilles2022metagraspnet}. Many works constrain their data to top-down grasps with oriented rectangles labels \cite{lenz2015deep, depierre2018jacquard, vuong2023graspanything}, but the recent trend is mainly towards object-centric grasps poses \cite{eppner2023abw2g, gilles2022metagraspnet, eppner2021acronym}. In that paradigm, ACRONYM \cite{eppner2021acronym} has recently been leveraged to train advanced learning architectures \cite{urain2023se3diffusionfield, barad2023graspldm, chen2024nsgf}. This success suggests that large and diverse enough datasets lead to great advances in the field.

%--------------------------------------------%
% Quality Diversity
%--------------------------------------------%

\textbf{\textit{Quality Diversity.}} 
Quality-diversity methods are optimization algorithms that aim to generate a set of diverse and high-performing solutions to a given problem \cite{cully2022qd}. QD has successfully been used to produce grasping datasets \cite{huber2023quality, huber2024speeding} that can be used in the real world by leveraging dedicated quality criteria \cite{huber2023domainrandomization}. The QDG-6DoF framework, dedicated to the generation of object-centric grasp poses, appeared to be significantly much more sample efficient than state-of-the-art approaches \cite{huber2024speeding}. The present work extends QDG-6DoF to data augmentation techniques, proposing a framework for the rapid creation of large-scale grasping datasets.

%--------------------------------------------%
% Object augmentation
%--------------------------------------------%

\textbf{\textit{Data augmentation.}} Most of the data augmentation methods used in the learning to grasp literature concerns images \cite{mahler2017dexnet2}\cite{van2018learning}. 
While augmentations of 3D data are largely studied in computer vision \cite{Zhu2024AdvancementsInPointCloudData}, applications for robotic grasping interactions remain nascent. Recent works focus on the approaching part of grasping by processing temporal data \cite{yamamoto2024real} or applying rigid transforms to grasping trajectories \cite{helenon2023fsta}. The present paper augments a dataset of object-centric grasps by applying deformations to the object meshes and transferring previously found grasps to the augmented objects.

%% file: tex_files/3_method.tex
%%%%%%%%%%%%%%%%%%%%%%%%%%%%%%%%%%%%%%%%%%%%%%%%%%%%%%%%%%%%%%%%%%%%%%%
%                               Method
%%%%%%%%%%%%%%%%%%%%%%%%%%%%%%%%%%%%%%%%%%%%%%%%%%%%%%%%%%%%%%%%%%%%%%%

\section{METHOD}

\input{tex_files/figures/qdgset_aug_principle}

%--------------------------------------------%
% Quality-diversity
%--------------------------------------------%

\textbf{\textit{Quality-diversity.}} Let $\mathcal{B} \subseteq \mathbb{R}^{n_b}$ be the \textit{behavior space} of dimension $n_b$, $\Theta$ the parameter space, and $\phi_{\mathcal{B}}:\Theta \rightarrow \mathcal{B}$ the \textit{behavior function}, which assigns a \textit{behavior descriptor} $b_\theta = \phi_{\mathcal{B}}(\theta)$ to each solution $\theta$. The \textit{fitness function} is $f:\Theta\rightarrow \mathbb{R}$, and $d_{\mathcal{B}}:\mathcal{B}^2 \rightarrow \mathbb{R}$ is a distance function within $\mathcal{B}$. The goal of QD optimization is to generate an \textit{archive} $A$ such that:
\begin{equation*}
\left\{\begin{matrix}
\forall b \in \mathcal{B}_{reach}, \, \exists \theta \in A, \, d_{\mathcal{B}}(\phi_{\mathcal{B}}(\theta), b) < \epsilon \\
\forall \theta' \in A, \, \theta' =  \text{argmax}_{\theta\in N(b_{\theta'})}f(\theta) 
\end{matrix}\right.
\label{eq:qd_background}
\end{equation*}
where $\mathcal{B}_{reach} \subseteq \mathcal{B}$ is the \textit{reachable behavior space}, $\epsilon\in\mathbb{R}^{+*}$ defines the density of $\mathcal{B}_{reach}$ paving, and $N(b_{\theta'})= \{ \theta \mid neighbor_{d_{\mathcal{B}}}(b_\theta, b_{\theta'}) \}$ is the set of solutions with close projections in $\mathcal{B}$. $\phi_{\mathcal{B}}$ is deterministic.

%I: nb is not defined, \theta not defined. Why is it 2?

% because it is d(b_theta_1, b_theta_2) so BxB

%-------------------------------------------%
% Object-centric grasps
%--------------------------------------------%

\textbf{\textit{Object-centric grasps.}} Let $g\in SE(3) \times \mathbb{R}^n$ be a grasp, with $n$ being the number of internal Degrees-of-Freedom (DoF). Similarly to many works on object-centric grasps for 2-finger grippers \cite{newbury2023review}, we here assume the gripper to be wide open at first and closed as much as possible for performing the grasp. The grasp is thus represented as a 6-DoF pose relative to the object frame, such that $g\in SE(3)$.

%--------------------------------------------%
% Data generation
%--------------------------------------------%

\textbf{\textit{Data generation.}} QDG-6DoF is a framework that exploit QD for generating 6DoF poses \cite{huber2024speeding}. Taking the 3D models of a gripper and an object as input, this framework explores their interactions in simulation to find grasp poses. Commonly used priors and QD optimization are used to drive the sampling of grasp poses. This process appeared to be significantly more sample efficient than approaches used to build previous object-centric datasets \cite{eppner2021acronym,eppner2023abw2g}.

In QDG-6DoF, grasping positions are sampled based on expert priors, such as the gripper always being close to the object surface to make the search more efficient. Dedicated QD mutation-selection mechanisms are then leveraged to drive the exploration of the search space toward the regions that are the most likely to produce new successes. The result is a large set of object-centric grasp poses all over the object surface. Our proposal extends QDG-6DoF: the chosen parameters structuring the core QD optimization process – the above introduced $\Theta$, $\mathcal{B}$, $d_{\mathcal{B}}$, $\phi_{\mathcal{B}}$, $f$ and $A$ – are the same as the vanilla QDG-6DoF \cite{huber2024speeding}. While QDG-6DoF works with many kind of grippers, we here only consider a Franka Emika Panda 2-finger one as it was used in the widely leveraged ACRONYM set \cite{eppner2021acronym}.

The generation process of QDG-6DoF must be started from scratch for any new object. This limitation alleviates the capabilities to scale-up the production of data. Yet, previous optimizations should be leveraged for similar objects. %\hl{: a set of grasp poses produced for a given mug is likely to be a good starting point to another mug, or even a cup.}
%IGN: In case of needing to remove things to make everything fit, reviewer suggests removing this highlighted sentence

% J: 100% agree 

%--------------------------------------------%
% Transfer learning
%--------------------------------------------%

\textbf{\textit{Transfer learning.}} Leveraging previous training to avoid starting from scratch is called \textit{transfer learning}. Commonly used in machine learning \cite{vision-tactile2023}, recent works explore this idea for QD \cite{salehi2022few}.
The present work proposes a simple approach for scaling up the production of grasping repertoires with QDG-6DoF. A bootstrapping archive of successful grasps $A_s^b$ is generated for a given object in a standard QDG-6DoF optimization. This archive is then used as initial population for the generation of grasps on a new-but-similar object. All the elements from $A_s^b$ are evaluated on the new object at the initialization generation of the evolutionary process. The optimization can thus be pursued to refine the outcome archive if necessary.

%I: explain what it contains

%--------------------------------------------%
% Data augmentation
%--------------------------------------------%

% repeated too many time:
%The present work extends the QDG-6DoF framework to scale up the production of large and diverse grasping datasets.

% to conclude:
%The previously described transfer learning approach accelerates the generation of diverse grasps on similar objects, as long as the objects are similar enough for some grasp poses to be successfully transferred (section \ref{sec:result_and_discussion}).

\textbf{\textit{Data augmentation.}} A first dataset of grasp poses is generated on a large set of core objects. Let $\textsc{P} = (\mathcal{P}_i)_{1 \leq i \leq n}$ be the dataset consisting of $n$ initial objects $\mathcal{P}_i$. Each initial object $\mathcal{P}_i$ of $\textsc{P}$ is defined by the following set of points:
\begin{center}
    $ \mathcal{P}_i = \left\{ p_j \in \mathbb{R}^3 \mid p_j = (x_j, y_j, z_j), \, j = 1, 2, \dots, m_i \right\}$
\end{center}
where $m_i$ represents the number of points in the point cloud of $ \mathcal{P}_i$. An object augmentation $\mathcal{P}_{i}^a$ of $\mathcal{P}_i$ is defined such that:
\begin{center}
    $\mathcal{P}_{i}^a = \left\{  \varphi_{D_a}(p_j) = D_a \cdot p_j  \mid p_j \in \mathcal{P}_i \right\}$.
\end{center}
where $D_a = \textsc{diag}(\alpha_1, \alpha_2, \alpha_3) \mid (\alpha_1, \alpha_2, \alpha_3) \in \mathbb{R}^3$ and the function $\varphi_{D_a}$ is defined as follows:
\begin{center}
\begin{align*}
\varphi_{D_a} :& \quad \mathbb{R}^3  \hskip0.8em \to \hskip0.8em \mathbb{R}^3 \\ 
&\begin{pmatrix}
x_j \\
y_j \\
z_j
\end{pmatrix}
\mapsto 
\underbrace{\begin{pmatrix}
\alpha_1 & 0 & 0 \\
0 & \alpha_2 & 0 \\
0 & 0 & \alpha_3 
\end{pmatrix}}_{D_a}
\begin{pmatrix}
x_j \\
y_j \\
z_j  
\end{pmatrix}
\end{align*}
\end{center}

The deformation applied by $\varphi_{D_a}$ to each point in $\mathcal{P}_i$ is characterized by the matrix $D_a$, where each diagonal element represents the scaling factor in the respective principal directions $x$, $y$, and $z$ of the object, centered at zero. The coefficients of $D_a$ are randomly chosen from an empirically defined range $[0.5, 1.5]$, which physically constrains the scale of all deformations within a range that makes most of the objects graspable by the considered gripper.
The augmented object dataset resulting from this method is noted $\textsc{P}' = (\mathcal{P}_{i}^{a}){}_{1 \leq i \leq n , 1 \leq a \leq a_{\text{max}}}$, where $a_{max}$ is the number of augmentations. $\textsc{P}'$ owns a new but similar object to the original $P$ database, on which many grasps can be transferred. %The combination of new objects from $\textsc{P}'$ with the bootstrapping of previously generated grasps on  $\textsc{P}$ allowed to scale up the dataset by an order of magnitude.

%IGN: How was chosen the defined range? Empirically?
%IGN: How do you know a_max?

%Then, each core object was submitted to a specific random perturbation along each of its frame axes. This resulted into a new but similar object, on which many grasps could be transferred. The combination of the transformation of core object with the bootstrapping of previously generated grasps allowed to scale up the dataset by an order of magnitude.

\textbf{\textit{Grasp transfer.}} Fig. \ref{fig:qdgset_augmentation_principle} overviews the proposed approach. A 3D model of a gripper and an object contact meshes are provided to QDG-6DoF for generating some grasp poses. The 3D model of the object is then augmented by applying the previously described linear deformations. The resulting augmented objects are then provided with a QDG-6DoF process, which is initialized with the grasps found on the reference object. The optimization process is stopped right after this bootstrapping step, reducing the number of evaluations. It results in new objects and associated grasp poses, obtained at limited cost (see section \ref{sec:result_and_discussion}).

\textbf{\textit{Objects.}} Objects models from 6 publicly available datasets are included in QDGset: KIT \cite{kasper2012kit}, 3DNet \cite{wohlkinger20123dnet}, YCB \cite{calli2015benchmarking}, GraspNet-1Billion \cite{fang2020graspnet}, EGAD \cite{morrison2020egad} and ShapeNet \cite{chang2015shapenet}. Some of those objects were already used in grasping papers \cite{eppner2021acronym,fang2020graspnet}, sometimes combining two sets \cite{mahler2017dexnet2}. But none of them gathered all of them in a unique, large-scale grasping database. Those sets include primitive, daily, complex, and adversarial objects. This allows to explore the grasping capabilities of a given gripper on a wide range of targets. This diversity is even higher than previous datasets that allowed to learn generalizing grasping policies \cite{eppner2021acronym}. It is worth to note that some datasets are far larger than others (Fig. \ref{fig:qdgset_objects_description}). Small datasets were also added due to their popularity \cite{kasper2012kit, wohlkinger20123dnet}, the access to real objects \cite{calli2015benchmarking}, or the capability to 3D print them \cite{morrison2020egad}.

Fig. \ref{fig:object_size_histplot} displays the distribution of object sizes in QDGset. On the left are plotted the sizes of objects from the included databases. While all the object sizes are smaller than 0.25m, ShapeNet object sizes lie between 0.6 m and 1 m. Such large sizes would prevent the considered gripper to grasp most of them. The ShapeNet objects were thus resized to make them close to daily objects that can be grasped by a human hand. To do so, a scaling factor $\mu_i$ is applied to the point cloud of each object $\mathcal{P}_i$ such that the resulting shifted distribution is indeed close to the YCB's one (Fig. \ref{fig:object_size_histplot}, right).

%IGN: the notation of the scaling factor has not been included in subsection "data augmentation" when you speak about perturbation? Wasn't it defined before then?
% Before. the gripper was not presented. Is it your robotic setup? What happen if we choose other gripper, is the method depending on the gripper?

% J: Agree. The gripper is now introduced much earlier, as well as the answer to the generalization question.
% J : The mu parameter is not detailed in the paper as it is a minor point regarding the contribution. We'll put all the details online.

% NOTE : COMMENT EST CALCULÉE LE SCALING FACTOR POUR LE RESCALE ENTRE SHAPNET ET YCB? (ECHANTILLONNÉ D'UNE DISTRIB, STATIQUE...?)

Fast evaluation in simulation is critical and require light 3D models \cite{coumans2016pybullet}. Therefore, the object models with size greater than 300kB were downsampled to make them smaller than this weight threshold.

QDGset is meant to be easily scaled by the research community. Only 27\% of the available core objects have already been augmented. Plus, the proposed augmentation method could easily scale the number of objects to orders of magnitudes – only 10 augmentations were conducted on a small subset of objects.

%--------------------------------------------%
% QDGset
%--------------------------------------------%

\input{tex_files/figures/qdg_figure_meta_description}

\input{tex_files/figures/qdgset_object_size_histplots}

\input{tex_files/tables/comparison_6dof_grasp_datasets}

\input{tex_files/figures/qdgset_hist_nb_successes}

\textbf{\textit{QDGset.}} Table \ref{table:dataset_comparison} compares QDGSet with available 6DoF pose datasets for 2-finger grippers. The generated dataset contains 29651 objects, reaching 40353 objects when including the augmented ones. Considering the diversity of included objects (Fig. \ref{fig:vis_intro_qdgset}), augmented ones can be considered as new objects: an augmented bottle results in many realistic 3D meshes of bottles (Fig. \ref{fig:vis_augment_kit_5_22_111}). Their difference is comparable to mugs that can be found in KIT, 3DNet, or YCB datasets, or even several objects from redundant datasets such as KIT or ShapeNet. This amount of objects is 4.5 times greater than the widely used ACRONYM set \cite{eppner2021acronym}.

Moreover, QDGset contains 62M grasps, up to 3.5 times more than previous state-of-the-art \cite{eppner2021acronym}. Both objects and grasp poses are diverse (Fig. \ref{fig:vis_intro_qdgset}) and have been produced with QDG-6DoF sample efficiency \cite{huber2024speeding}. This work takes a step further with the proposed bootstrapping method, which increases the sample efficiency even more.

%------------------- J : ?
The number of samples was not provided in table \ref{table:dataset_comparison} – it is not meaningful in the case of object-centric-grasp labels, as such data is meant to be adapted to the format of the targeted architecture. While the number of grasp poses cannot be increased without collecting new data for Graspnet-1Billion, QDGSet grasps can be augmented with as many samples as can be created from the provided data – by taking different camera viewpoints in particular.
%-------------------

Fig. \ref{fig:qdgset_nb_grasp_per_object} shows the number of grasps per object in QDGset. Most of the objects yield from 1000 to 5000 grasp poses. Those distributions reflect the complexity of the 3D models: the average number of grasps per KIT and 3DNet object are respectively 3391 and 3787, while it reaches 1286 for ShapeNet. No grasp was found for some of the YCB, GraspNet, and ShapeNet objects. This is primarily due to the limitation of the Franka Emika Panda gripper, mainly because of its size. Thus, QDGset contains not only information on the capabilities but also on the gripper limitations. 

%IGN: we speak about this gripper again without saying that is the one that we will be using. It is needed to introduce the gripper and specify why we are using this one

% J: Done, at the end of the "Data generation" subsection.

%% file: tex_files/figures/qdgset_aug_principle.tex
\begin{figure}[t]
  \centering
\centering
  \includegraphics[width=\columnwidth]{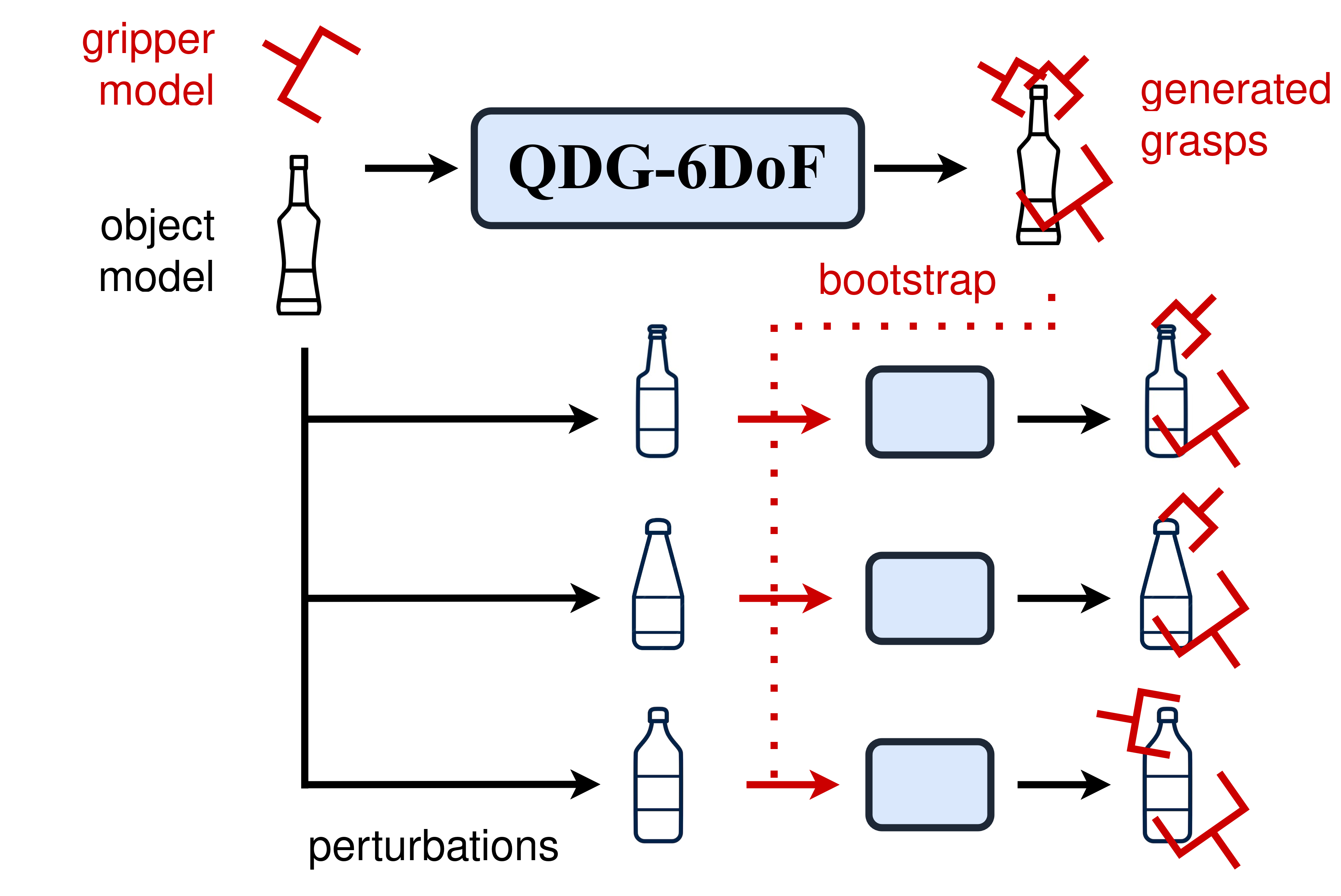}
  \caption{\textbf{Data augmentation principle.} A gripper and an object model are provided to QDG6DoF \cite{huber2024speeding} to generate grasp poses. The object model is then perturbed to generate new objects leveraged in new QDG6DoF runs. Each of those runs are bootstrapped with previously found grasps.}
  \label{fig:qdgset_augmentation_principle}
\end{figure}

%IGN: supervisor suggests using "variations" instead of "perturbations" in both description and figure

%% file: tex_files/figures/qdg_figure_meta_description.tex
\begin{figure}[t]
  \centering
\centering
  \includegraphics[width=0.90\columnwidth]{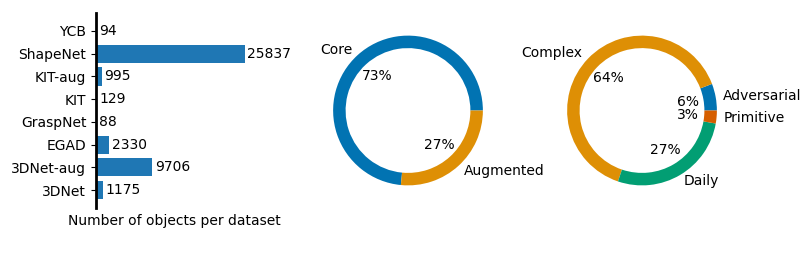}
  \caption{\textbf{QDGset objects.} (Left) Number of objects per dataset; (center) ratio of augmented objects; (right) object categories.}
  \label{fig:qdgset_objects_description}
\end{figure}

%IGN: you have to explain differences and features of complex objects vs daily vs adversarial vs primitive (like shape, geometry, articulated, deformable...)

% J: This will be detailed online in the appendices. Unfortunately, 6 pages is short.

%% file: tex_files/figures/qdgset_object_size_histplots.tex
\begin{figure}[t]
  \centering
\centering
  \includegraphics[width=0.95\columnwidth]{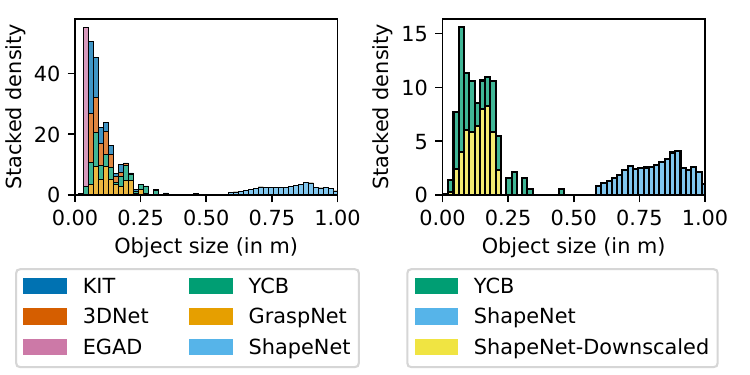}
  \caption{\textbf{Object sizes.} (Left) QDGset object size distribution before ShapeNet rescale; (right) shapenet rescale based on YCB sizes.}
  \label{fig:object_size_histplot}
\end{figure}

%% file: tex_files/tables/comparison_6dof_grasp_datasets.tex
\begin{table*}[t]
\centering
\begin{tabular}{ ||c || c | c | c | c | c | c | c | c  ||}

\hline

\makecell{Dataset} & \makecell{Data\\Representation}  & \makecell{Grasps\\per object} & \makecell{Num.\\objects} & \makecell{Num.\\grasps} & \makecell{Num.\\samples} & \makecell{Data\\type} & \makecell{Grasp\\sampling} & \makecell{Quality\\annotation}   \\

\hhline{||=||=|=|=|=|=|=|=|=||}

VR-Grasping-101 \cite{yan2018learning} & Image & 10-20 & 101 & 151K & - & Sim & Augmented demonstrations & Trial-and-error \\
\hline
GraspNet-1Billion \cite{fang2020graspnet} & Image & 3-9M & 88 & 1.2B & 97K & Real & Antipodal + uniform & Analysis \\ % Antipodal-based (uniformly on a sphere)
\hline
Kappler et al. \cite{kappler2015leveraging} & Point Cloud  & 500 & 80 & 300K & 700 & Sim & Approach + uniform & Analysis \\ % Approach-based (uniformly on bounding box)
\hline
6DOF Graspnet \cite{mousavian20196} & Point Cloud  & 34K & 206 & 7M & 206 & Sim & Approach + random & Sim (FleX) \\
\hline
Eppner et al. \cite{eppner2023abw2g} & Point Cloud & 47.8M & 21 & 1B & 21 & Sim & SE(3) uniform & Sim (FleX) \\
\hline
EGAD! \cite{morrison2020egad} & Point Cloud & 100 & 2231 & 233K & 2231 & Sim & Antipodal + uniform & Analysis \\
\hline
ACRONYM \cite{eppner2021acronym} & Point Cloud & 2K & 8872 & 17.7M & - & Sim & Antipodal + random & Sim (FleX) \\
\hline
MetaGraspNet \cite{gilles2022metagraspnet} & Point Cloud & 5K & 82 & - & 217 & Sim & Antipodal + uniform & Sim (FleX) \\

\hhline{||=||=|=|=|=|=|=|=|=||}

QDGSet\textit{-core} & Point Cloud &  \textbf{1-5K} & \textbf{29651} & \textbf{43M} & - & Sim & QDG-6DoF & QD-DR \\
\hline
QDGSet & Point Cloud & \textbf{1-5K} & \textbf{40353} & \textbf{62M} & - & Sim & QDG-6DoF & QD-DR \\

\hline

\end{tabular}
\caption{\textbf{Comparison of QDGset with available object-centric grasping datasets.}}
\label{table:dataset_comparison}
\end{table*}

%IGN: what does "-" mean? Like in between those numbers?
%J: Exactly; it comes from [49] I think

%% file: tex_files/figures/qdgset_hist_nb_successes.tex
\begin{figure}[t]
  \centering
\centering
  \includegraphics[width=\columnwidth]{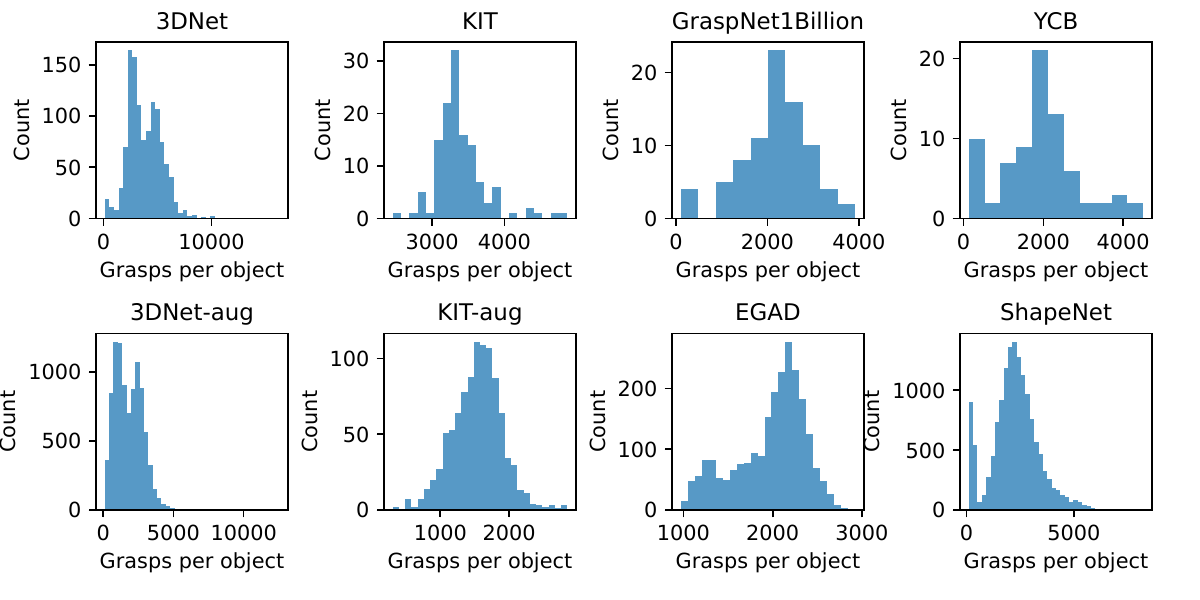}
  \caption{\textbf{Number of successful grasps per object.} Most of the distributions lie between 1000 and 5000 grasps per objects. Those distributions reflect the complexity of the subset of objects.}
  \label{fig:qdgset_nb_grasp_per_object}
\end{figure}

%% file: tex_files/4_experiments.tex
%%%%%%%%%%%%%%%%%%%%%%%%%%%%%%%%%%%%%%%%%%%%%%%%%%%%%%%%%%%%%%%%%%%%%%%
%                              Experiments
%%%%%%%%%%%%%%%%%%%%%%%%%%%%%%%%%%%%%%%%%%%%%%%%%%%%%%%%%%%%%%%%%%%%%%%

\input{tex_files/figures/visualization_augmented_object_subsample_kit}

\section{EXPERIMENTS}   

%-------------------------------------------%
% Bootstrapping QD
%-------------------------------------------%

\textbf{\textit{Bootstrapping QD.}} To evaluate the bootstrapping approach, a set of experiments is conducted on several augmented objects. First, a success archive $A_s^b$ of diverse grasp poses is generated for a set of reference objects. Variations are then applied to the objects to produce new ones. For each resulting object, all the grasps from the bootstrapping archive $A_s^b$ are evaluated once to initialize the QD optimization. The evolutionary process is then carried out until a maximum number of evaluations is reached. 

The reference objects were chosen from two of the most popular object datasets, KIT and 3DNet, to express the diversity of QDGset objects (Fig. \ref{fig:vis_augment_kit_5_22_111}): the KIT n°5, 22, 111, and the 3DNet n° 22, 150, and 410. Between 50 and 100 augmented objects were produced. The number of robust grasps generated throughout the search is then studied by comparing the performance of QDG-6DoF started from scratch with optimizations that leverage the proposed transfer learning approach. QDG hyperparameters are similar to the vanilla paper \cite{huber2024speeding}. The grasping repertoires were generated on 100k trials. A grasp is considered robust if it resists to three trials under domain randomization constraints \cite{huber2023domainrandomization}. Unlike in the generation of QDGset, the QD optimization is not stopped after the bootstrapping process.

%IGN: Remove that part since in the image there are only KIT objects

\textbf{\textit{Grasp transfer.}} The efficiency of the proposed transfer learning approach has thus been evaluated by analyzing the transferability rate between grasps generated on 3DNet and KIT databases to the 3DNet-aug and KIT-aug augmented ones. Each of them consists on about 10 augmented objects for each reference model. The resulting quality distributions are then compared to the core ones.

The hyperparameters and the non-aggregated results can be found in the appendices shared on the project webpage\footnote{\url{https://qdgrasp.github.io/}}.

%% file: tex_files/figures/visualization_augmented_object_subsample_kit.tex
\begin{figure}[t]
  \centering
\centering
  \includegraphics[width=0.95\columnwidth]{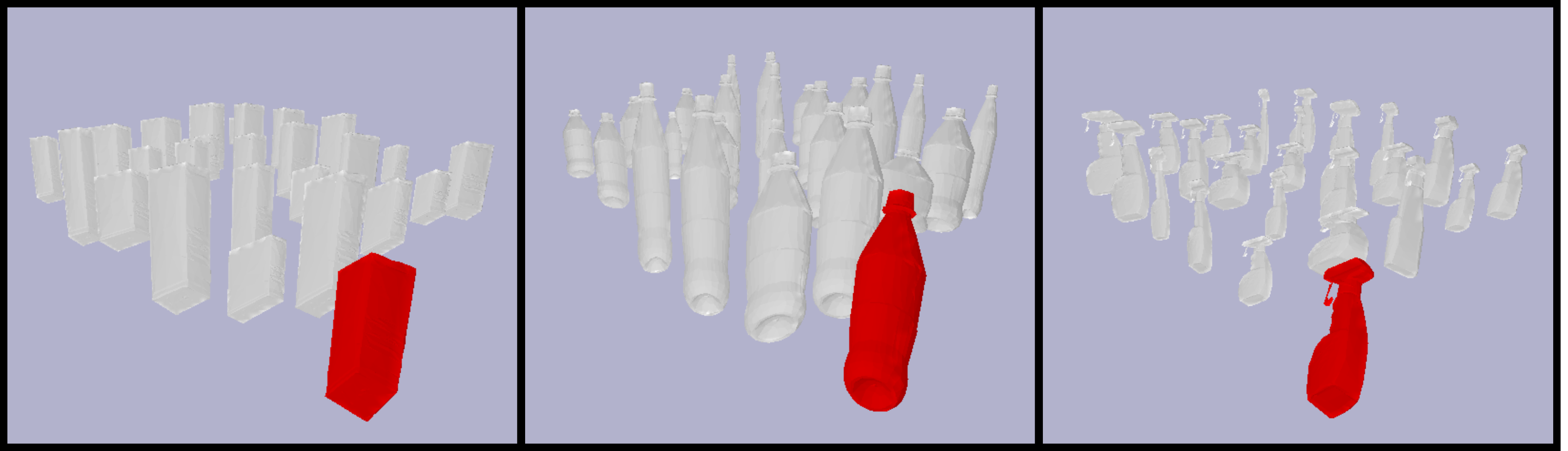}
  \caption{\textbf{3D meshes augmentation.} In red are the reference objects (KIT n°5, n°22, n°111), and in grey a subsample of 25 resulting augmentations.}
  \label{fig:vis_augment_kit_5_22_111}
\end{figure}

%% file: tex_files/5_results_and_discussion.tex
%%%%%%%%%%%%%%%%%%%%%%%%%%%%%%%%%%%%%%%%%%%%%%%%%%%%%%%%%%%%%%%%%%%%%%%
%                        Results and discussion
%%%%%%%%%%%%%%%%%%%%%%%%%%%%%%%%%%%%%%%%%%%%%%%%%%%%%%%%%%%%%%%%%%%%%%%

\section{RESULTS AND DISCUSSION}
\label{sec:result_and_discussion}

\input{tex_files/figures/bootstrap_exp_sample_efficiency}

\input{tex_files/figures/qdgset_fit_transfer_rate_analysis}

\input{tex_files/figures/vis_bootstrap_heatmap}

%-------------------------------------------%
% [RESULTS] Bootstrapping QD
%--------------------------------------------%

\textbf{\textit{Bootstrapping QD.}} Fig. \ref{fig:bootstrap_exp_sample_efficiency} shows the obtained results for the evaluation of the bootstrapping approach on the grasp generation. The box plot shows the differences in generated grasps between bootstrapped runs and from-scratch optimization. Overall, 76\% over the 431 runs lead to more robust grasps when stopping after the bootstrapping step. The obtained data have been aggregated on the boxplot into reference objects that lead to a significant improvement, and those for which the difference is in the error margin. Among the tested reference objects, 4 of them lead to at least 75\% of positive difference (\textit{Better}). The interquartile range of the last one overlaps the null line (3DNet n°22), leading to no gain nor cost (\textit{Constant}). In this study, bootstrapping the QD optimization process results in a beneficial – or, at worst, neutral – effect on the robust grasp generation.

When bootstrapping is beneficial (3DNet n°410, KIT n°5, 22, 111), and when stopping after initialization, the average number of robust generated grasps is 309 for about 5059 evaluations, leading to a ratio of 5059/309=16.4. From scratch, an average of 244 robust grasps is produced for the same number of evaluations, leading to a ratio of 20.7 evaluations per grasp. Therefore, this approach reduces the required per grasp evaluations number by up to 20\%.

However, the difference after 100k evaluations is close to zero and often lower (Fig.\ref{fig:bootstrap_exp_sample_efficiency}, left). The bootstrapping has thus a detrimental effect on the long run. Transferring previously found grasps imposes strong constraints on the way to explore the search space to find solutions: if the objects are not similar enough, the grasps that are likely to evolve into new grasp poses on one object can stick the search to a local minimum \cite{huber2023quality}. The line plots show the learning curves of good bootstrapping runs (center) and neutral bootstrapping runs (right). Those plots provide the learning dynamics that are condensed in the boxplots: 1) stopping after the bootstrapping evaluations leads to a positive or neutral effect, and 2) continuing the optimization process would be detrimental.

%IGN: I can't see this fact in figure 7. If can't see remove this sentence.
% J: Left figure, in orange. close to 0 for "better", lower for "constant". 

The dataset has thus been generated through scratch optimization for all the reference objects \cite{huber2024speeding} and with the bootstrapping approach for all the augmented objects. The optimization process has been stopped right after the bootstrapping evaluations.

%-------------------------------------------%
% [RESULTS] Grasp transfer
%--------------------------------------------%

\textbf{\textit{Grasp transfer.}} Fig. \ref{fig:transfer_rate_fit_distrib_3dnet_kit} shows results obtained by generating the augmented datasets with the bootstrap approach. For 78\% of the 1304 considered reference objects, 10 out of 10 transfers lead to at least 500 successful grasps (left). Moreover, the distributions of grasp qualities are similar for the reference grasp sets and the augmented ones (center and right). This shows that the proposed bootstrapping method leads to successful transfer from core objects to augmented ones while maintaining the same quality distribution. An example of generated grasps for a given core object and 3 related augmentations is provided in Fig. \ref{fig:bootstrap_heatmap_example}. A large part of the core archive has successfully been transferred, leading to new grasping archives that cover the whole object surface. Furthermore, the resulting grasps are of both good and bad qualities. Having various grasp quality is a desirable feature for learning techniques relying on negative samples \cite{eppner2021acronym}.

%-------------------------------------------%
% [DISCUSSION] Objects
%--------------------------------------------%

\textbf{\textit{Core objects.}} The objects included in the dataset are critical to get results through learning. Some of the included objects are realistic (e.g. KIT, YCB, ShapeNet), but the necessary rescale step somehow orients the dataset toward some cases. For example, ShapeNet contains many models of large objects, like vehicles. Rescaling them created a range of small objects that are similar to little toys - which is likely to lead to better performance in related industrial scenarios. On the contrary, EGAD contains adversarial objects which are not likely to be found in the real world. Finding a good balance between realistic and adversarial objects should be a critical challenge to get the best out of the proposed approach. Nevertheless, the ease of producing data for a new kind of object suggests that this balance can finely be controlled by practitioners. For example, subsets of QDGset can be used for a specific purpose (e.g. grasp pans from a convey belt in an industrial scenario).

%IGN: Supervisor tells to end sentence on "specific purpose", like removing highlighted text
% J: I remember a reviewer feedback about those kind of statement, without concrete examples...I guess it is a minor detail.

%-------------------------------------------%
% [DISCUSSION] Augmented objects
%--------------------------------------------%

\textbf{\textit{Augmented objects.}} Similarly, the range of variations considered for the augmentation should be adapted to the targeted domains. Let's consider a scenario in which the robot only has to target objects that are larger than its gripper. A learning model should absorb data that contains information on how to grasp subparts of large objects. If the range of perturbation is too small, the core objects are downscaled, resulting in a set of grasps that catch small objects that fit into the gripper, such that most of the grasps consist of aligning the center of the gripper with the object center of mass. This is likely to strongly deteriorate the quality of a resulting trained model for the considered task. A well-suited range of variations would transform objects to make them as close to the targeted ones as possible.

%IGN: supervisor suggests using "variations" instead of "perturbations" 
%J: done

The objects could also be augmented with other strategies, such as local perturbation, noise, crossover between objects, or different physical properties. Similarly, the bootstrap process can be initiated with more advanced approaches. With such a large and diverse database, a model could be trained to propose bootstrapping candidates. While any 6-DoF grasp sampler that is not restricted to a single viewpoint can be a good starting point \cite{urain2023se3diffusionfield}, the model can be tuned not only to produce good grasps but also to propose grasps that are likely to speed up the coverage of the outcome archive \cite{salehi2022few}.

%-------------------------------------------%
% [DISCUSSION] Limits and perspectives
%--------------------------------------------%

\textbf{\textit{Limits and perspectives.}} QDG-6DoF allows to generate multi-finger object-centric grasps too \cite{huber2024speeding}. Extending the database to grippers beyond parallel-jaw ones is a promising perspective. This raises the question of the right learning architecture for predicting grasp poses beyond 2-finger ones. Some recent works explore this idea \cite{li2022survey, geograsp2019, geograspevo2023}, but robust results on real platforms are yet to be demonstrated.

Similarly, the transfer learning method could easily be extended to new grippers \cite{jaquier2023transfer}. Instead of using a bootstrapping archive $A_s^b$ of a pair (object 1, gripper 1) to bootstrap a pair (object 2, gripper 1), $A_s^b$ can be used for (object 1, gripper 2), or even (object 2, gripper 2). This is likely to work on similar grippers, for example, two variants of parallel-jaw ones, but it might struggle if the grippers or the objects are too different from the bootstrapping ones. 

A key feature of the generated grasps is their diversity. QDGset contains diverse grasps on diverse objects. Associating the grasps with semantic labels opens many promising perspectives regarding affordances \cite{kokic2017affordance, mandikal2021learning}. For example, QDGset contains many ways of grasping handles – allowing several manipulation tasks like pouring, or placing. The advent of large language models \cite{zhao2023survey} brings a powerful tool to process semantic data to include the low-level skill of grasping into a high-level robot decision process. Extending QDGset to semantic labels is considered for future work.

%IGN: remove that part. It is off topic.
%J: There is like a revolution right now with LLMs, and the capabilities they provides to almost any problem. Here their usage on QDGset is so obvious and straightforward, that not mentioning it as perspective would mean that we have no idea of how interesting our work. At least 3 different sub-groups in the lab are working on that right now, even before we finished this paper... LLMs + robotics yields a survey every week since the release of chatGPT ...

Lastly, the proposed method relies on previous works on QD for grasping for sim2real transfer purpose \cite{huber2023quality, huber2024speeding}. Leveraging QDGset for learning generalizing grasping policies for real robotics is considered for future work.

%% file: tex_files/figures/bootstrap_exp_sample_efficiency.tex
\begin{figure}[t]
  \centering
\centering
  \includegraphics[width=0.95\columnwidth]{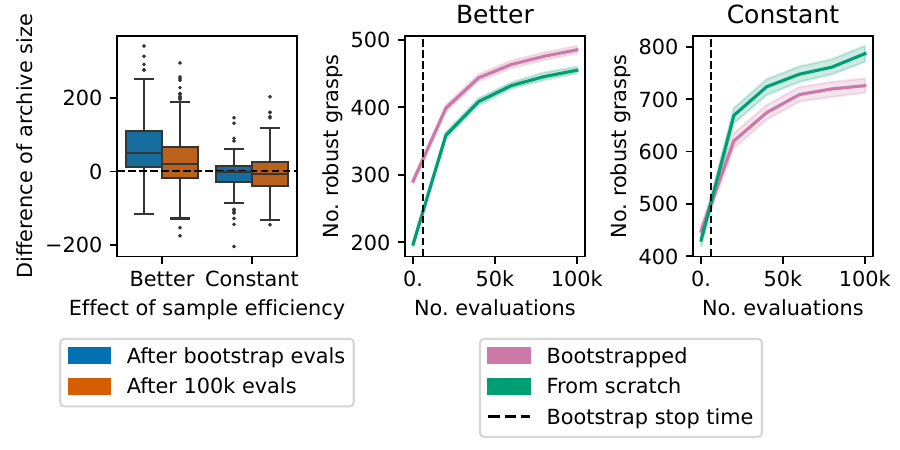}
  \caption{\textbf{Sample efficiency evaluation.} (Left) Difference of number of generated grasps when stopping at bootstrap time, or on the long run; (center) aggregated results for good bootstrap effect; (right) aggregated results for null bootstrap effect.}
  \label{fig:bootstrap_exp_sample_efficiency}
\end{figure}

%IGN: Change "evals" in left graph by "trials". I think Stephane asked to remove this image (?)

% J: we can set "eval" in the camera ready version.
% J: about Stephane comment, he is right that this figure is now easy to read. But this is how we aim to buy the validation from the reviewers. After that the paper will only be about the database and the method, not those curves. I am sure he would understand it if we would talk to him about it.

%% file: tex_files/figures/qdgset_fit_transfer_rate_analysis.tex
\begin{figure}[t]
  \centering
\centering
  \includegraphics[width=0.95\columnwidth]{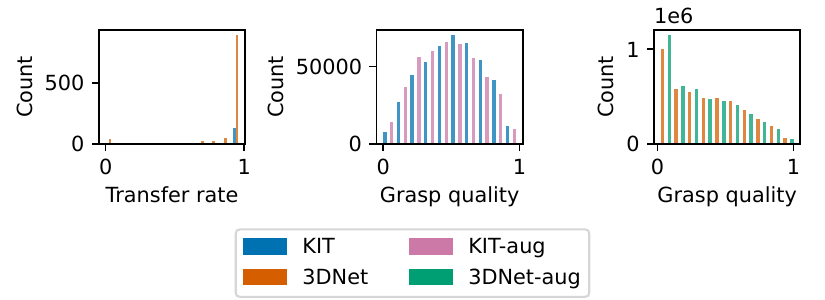}
  \caption{\textbf{Transferability.} (Left) Transfer rate from core object to augmented one - successful if $\leq$500 grasps are transferred; (center) grasp quality distribution on KIT-* data; (right) same on 3DNet-* data.}
  \label{fig:transfer_rate_fit_distrib_3dnet_kit}
\end{figure}

%% file: tex_files/figures/vis_bootstrap_heatmap.tex
\begin{figure}[t]
  \centering
\centering
  \includegraphics[width=0.95\columnwidth]{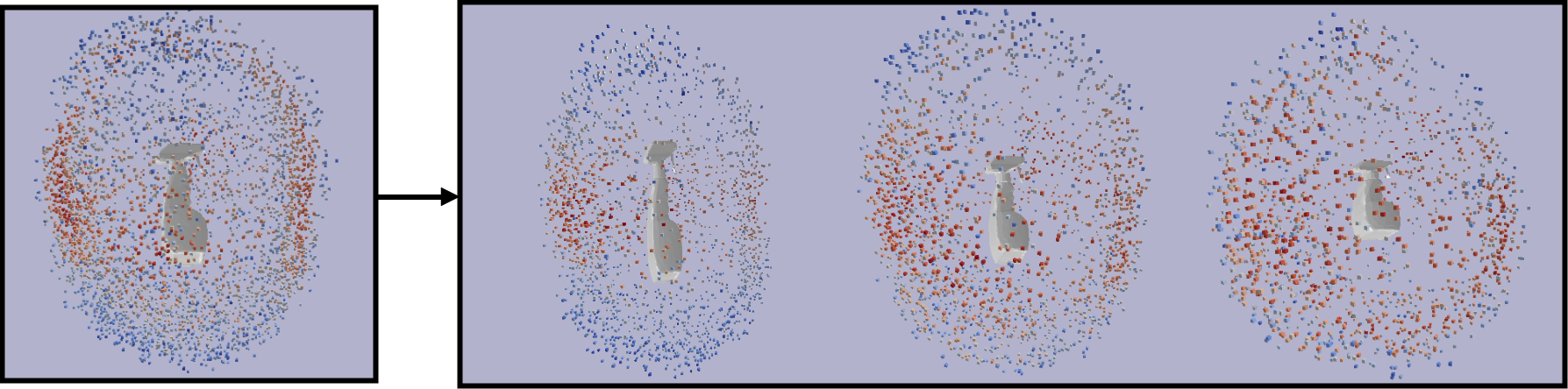}
  \caption{\textbf{Transferred grasping archive.} (Left) Example of a core object and the associated generated grasps (left) for KIT n°111, and the resulting grasping archives after transferring it on 3 augmented objects (right). Each dot corresponds to the end effector position for a given grasp pose. The color describes the quality, the hotter the better.}
  \label{fig:bootstrap_heatmap_example}
\end{figure}

%% file: tex_files/6_conclusions.tex
%%%%%%%%%%%%%%%%%%%%%%%%%%%%%%%%%%%%%%%%%%%%%%%%%%%%%%%%%%%%%%%%%%%%%%%
%                             Conclusions
%%%%%%%%%%%%%%%%%%%%%%%%%%%%%%%%%%%%%%%%%%%%%%%%%%%%%%%%%%%%%%%%%%%%%%%

\section{CONCLUSIONS}

This work introduces QDGset, a large-scale parallel jaw grasp dataset provided as object-centric poses. It was produced by extending QDG-6DoF to data augmentation for grasping. This approach combines the creation of new simulated objects from reference ones with the transfer of previously found grasps. This method can reduce the required number of evaluations for finding robust grasps by up to 20\%, while maintaining the same quality distributions. The obtained results demonstrate how QD can be used to generate new data for a particular grasping scenario. We believe that such a tool makes possible the gathering of a large collaborative dataset of simulated grasps which can be successfully used in the real world.

%IGN: Remove that part.

%% file: tex_files/acknowledgment.tex
%%%%%%%%%%%%%%%%%%%%%%%%%%%%%%%%%%%%%%%%%%%%%%%%%%%%%%%%%%%%%%%%%%%%%%%
%                           Acknowledgment
%%%%%%%%%%%%%%%%%%%%%%%%%%%%%%%%%%%%%%%%%%%%%%%%%%%%%%%%%%%%%%%%%%%%%%%

\section*{ACKNOWLEDGMENT}

This work was supported by the Sorbonne Center for Artificial Intelligence, the German Ministry of Education and Research (BMBF) (01IS21080), the French Agence Nationale de la Recherche (ANR) (ANR-21-FAI1-0004) (Learn2Grasp), the European Commission's Horizon Europe Framework Programme under grant No 101070381, by the European Union's Horizon Europe Framework Programme under grant agreement No 101070596, by Grant PID2021-122685OB-I00 funded by MICIU/AEI/10.13039/501100011033 and by the European Union NextGenerationEU/PRTR. This work used HPC resources from GENCI-IDRIS (Grant 20XX-AD011014320).